# A light-weight and efficient punctuation and word casing prediction model for on-device streaming ASR

*Jian You, Xiangfeng Li*

Cisco Systems - Shanghai, China

{jianyou, xiangfel}@cisco.com

**ABSTRACT**

Punctuation and word casing prediction are necessary for automatic speech recognition (ASR). With the popularity of on-device end-to-end streaming ASR systems, the on-device punctuation and word casing prediction become a necessity while we found little discussion on this. With the emergence of Transformer, Transformer based models have been explored for this scenario. However, Transformer based models are too large for on-device ASR systems. In this paper, we propose a light-weight and efficient model that jointly predicts punctuation and word casing in real time. The model is based on Convolutional Neural Network (CNN) and Bidirectional Long Short-Term Memory (BiLSTM). Experimental results on the IWSLT2011 test set show that the proposed model obtains 9% relative improvement compared to the best of non-Transformer models on overall $F_1$-score. Compared to the representative of Transformer based models, the proposed model achieves comparable results to the representative model while being only one-fortieth its size and 2.5 times faster in terms of inference time. It is suitable for on-device streaming ASR systems. Our code is publicly available.

## 1. INTRODUCTION

The automatic speech recognition (ASR) systems usually output transcripts without punctuation and word casing. This kind of unlabeled transcripts are unreadable and can easily cause misunderstandings, which makes ASR applications (e.g. Closed Caption) unusable. Therefore, it is essential to have a post-processing module to handle punctuation and word casing prediction. It can not only improve user experience but also benefit the possible subsequent NLP tasks (e.g. machine translation, dialogue system). With the advancement of ASR models, the on-device end-to-end streaming ASR becomes more and more popular [1][2][3][4]. As the post-processing module, the on-device punctuation and word casing prediction also become necessary. However, we found rare discussion on this on-device scenario.

In terms of feature type used by model, there are three types, i.e. lexical, acoustic, or a combination of both [5][6][7]. Due to easy access to large amount of well formatted text (e.g. public news website, Wikipedia website), lexical features have been widely adopted by punctuation and word casing prediction models. In this paper, we jointly train punctuation prediction and word casing prediction only based on lexical features.

With the emergence of Transformer, Transformer based models have been popularly employed in this scenario [8][9][10][11]. Hence, we classify the model structures into non-Transformer models and Transformer based models. For non-Transformer models, conditional random field (CRF) was used in earlier studies [12][13]. Convolutional Neural Network (CNN) was then used for punctuation prediction task [14][15][16]. Then the use of Long Short-Term Memory (LSTM) becomes popular [17][18][19][20]. For Transformer based models, the pre-trained encoder part of the Transformer was usually used (e.g. BERT [11], RoBERTa [10]). Due to the deep model depth of Transformer, the size of Transformer based models is usually very large, which impedes their deployment on edge devices.

In summary, our contributions are the following: 1) We propose a CNN-BiLSTM model to jointly predict punctuation and word casing only based on lexical features. To the best of our knowledge, this is the first work to concatenate Convolutional Neural Network (CNN) and Bidirectional Long Short-Term Memory (BiLSTM) modules for joint prediction of punctuation and word casing with only lexical features. 2) We well study the model performance (e.g. model size, inference time) for on-device requirements. 3) Experimental results on the IWSLT2011 test set show that the proposed model outperforms previous non-Transformer models on $F_1$-scores. Using a Transformer-based model for comparison, the proposed model not only achieves comparable results but also is one-fortieth the size and offers an inference time that is 2.5 times faster. 4) The source code of the model is publicly available[1].

We organize the paper as follows. In Section 2, we discuss the related work. Section 3 describes the proposed approach. Experimental details and results are provided in Section 4. We conclude the paper in Section 5.

## 2. RELATED WORK

In view of the effectiveness of CNN model in text classification tasks, CNN was used to predict punctuations in earlier studies. Che et al. [14] proposed two types of CNN model to predict punctuation, one views each word embedding as a vector, the other views each element in word embedding matrix independently. Żelasko et al. [15] used multiple layers of CNN model and they proposed a model consisting of multiple layers of BiLSTM for comparison. They concluded that CNN model yields higher precision while BiLSTM model has better recall. Augustyniak et al. [16] improved CNN model in Żelasko et al. [15] by retrofitting word embeddings.

---

[1]The code is available at https://github.com/frankyoujian/Edge-Punct-Casing

As longer context is helpful for better prediction performance, Recurrent Neural Network (RNN) based models have been employed for this prediction task. Tilk et al. [17] used Bidirectional Recurrent Neural Network (BRNN) with an attention mechanism on top of it. Tündik et al. [18] proposed a BiLSTM based model for punctuation prediction. Öktem et al. [19] proposed a parallel Gated Recurrent Units (GRU) based model. Although CNN and BiLSTM models have been extensively studied for this scenario independently, we have found no existing research that attempts to concatenate CNN and BiLSTM architectures for this specific prediction task.

Word casing plays the same important role as punctuation in sentence understanding. Several studies have verified that punctuation and word casing prediction are two tasks that could benefit from each other [20][21][22]. In this paper, we adhere to the same mechanism of multitask learning; however, we employ a distinct approach for processing the inputs of these two tasks, as explained in Section 3.

## 3. PROPOSED APPROACH

The proposed model architecture is showed in Figure 1. The inputs of the model are raw transcripts, which are usually the outputs of ASR systems. For each word in the transcripts, the model performs punctuation and word casing tasks simultaneously. For example, the input is "*will ai change our future*", the model outputs "*Will AI change our future?*".

In this paper, we consider four punctuation marks: comma (COMMA), period (PERIOD), question mark (QUESTION) and no punctuation (O). The punctuation is predicted to follow each word. For word casing, there are four types: all uppercase (UPP), capitalization case (CAP), mixed case (MIX) and all lowercase (O). The mixed case denotes cases like iPhone, PhD, etc.

To effectively extract information from embeddings of words, we use CNN based encoder layers. The inputs of encoder layers are the embeddings of subwords, which are the results of subword tokenizer. The encoder layer is based on a 1D convolution, which is followed by a RELU activation. The result of the activation is added to the original input, following the residual connection paradigm. Then layer normalization is performed on the sum. The CNN encoder layers here serve as language model in [23] while require fewer parameters to train and train faster compared to Transformer.

The sequence of CNN encoded subword embeddings $X = (x_1, ..., x_T)$ is then processed by several bidirectional LSTM layers which process the sequence in both forward and backward directions. It is worth noting that only the first subword will represent the whole word to be processed by BiLSTM layers while the subsequent subwords will be masked. The state $\vec{h}_t$ at time step $t$ of the forward recurrent layer is

$$\vec{h}_t = \phi(x_t, \vec{h}_{t-1}) \quad (1)$$

where $\phi$ is a recurrent activation function, here we use the long short-term memory units (LSTM). The backward recurrent layer calculates its state $\overleftarrow{h}_t$ at time step $t$ in a way similar to the forward recurrent layer, except that it processes the input $X$ in reverse order. Subsequently, the bidirectional state $h_t$ is formed by concatenating the forward layer state and backward layer state at time step $t$:

$$h_t = [\vec{h}_t, \overleftarrow{h}_t] \quad (2)$$

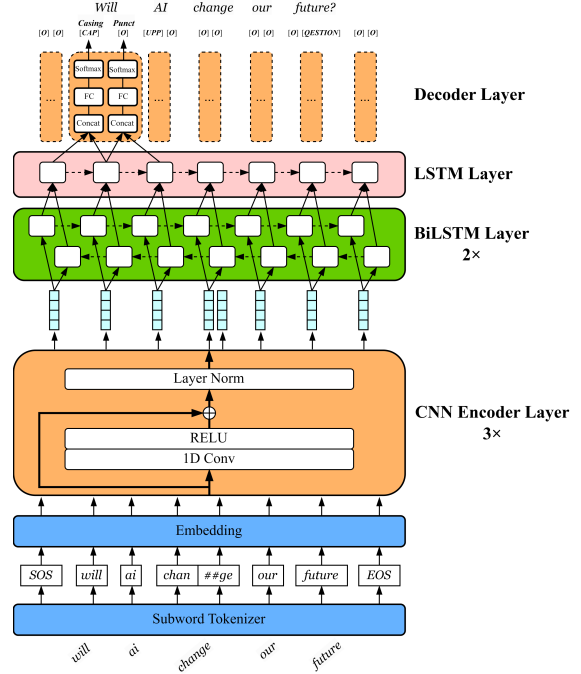

Figure 1: *The proposed model architecture.*

So this layer captures both past and future contexts for prediction in that the label of a word is considerably affected by the words following it, as it is by the words preceding it.

On top of the bidirectional layers, we use a unidirectional LSTM layer to sequentially process the bidirectional states and monitor the ongoing text position. The state of this layer at time step $t$ is

$$s_t = \phi(h_t, s_{t-1}) \quad (3)$$

The adjacent states are then concatenated as the input to the decoder layer. The output sequence of punctuation $(y^p_1,..., y^p_n)$ and casing $(y^c_1,..., y^c_n)$ can be derived from the sequence of state vectors $(z^p_1,..., z^p_n)$ and $(z^c_1,..., z^c_n)$ respectively, where $z^p_t = [s_t, s_{t+1}]$ and $z^c_t = [s_{t-1}, s_t]$, through transformations, as defined below:

$$y^p_t = softmax(W^p z^p_t + b^p) \quad (4)$$

$$y^c_t = softmax(W^c z^c_t + b^c) \quad (5)$$

The state $z^c_t$ is the concatenation of current and previous unidirectional state, since we think the previous token information is helpful to predict the current casing, e.g. if previous token punctuation is comma, then current casing is impossible to be leading capitalization. Contrarily, the state $z^p_t$ is the concatenation of current and next unidirectional state because the casing of next token does help to current punctuation prediction, e.g. if next token casing is leading capitalization, the current punctuation is most likely to be period. The concatenated state $z^p_t$ and $z^c_t$ are passed through two different branches which are both composed of full connected layer and softmax layer. In the above formulation, $W^p$, $b^p$ denote weights and bias of punctuation full connected layer and $W^c$, $b^c$ denote weights and bias of casing full connected layer. The output $(\boldsymbol{y}_t^p, \boldsymbol{y}_t^c)$ of the two branches are the predictions of punctuation and casing at time step $t$.

The learning objective of our model is to maximize the prediction performance of both punctuation and casing tasks. The parameters of CNN encoder layers, BiLSTM layers and LSTM layer are shared across both tasks. We train the shared parameters jointly based on the loss function defined over the outputs of both tasks. We compute the loss $\mathcal{L}^p$ for punctuation prediction task and $\mathcal{L}^c$ for casing prediction task through cross entropy loss function. The final loss $\mathcal{L}$ is a weighted average of $\mathcal{L}^p$ and $\mathcal{L}^c$:

$$\mathcal{L} = \mathcal{L}^c + \alpha \mathcal{L}^p \qquad (6)$$

In the above formulation, $\alpha$ is a predefined weight which is optimized to achieve the best prediction across both tasks.

## 4. EXPERIMENTS

### 4.1 Datasets

To ensure the diversity of training dataset, we collect public data from a variety of domains, e.g. public interview and podcast transcripts, Wikipedia public raw text and TED Talks transcripts. We clean original datasets by filtering out sentences containing punctuations that are not in (COMMA, PERIOD, QUESTION). Finally, we get about 300 MB texts that consist of 55M words. Table 1 shows the distribution of word counts as percentages across the sets.

| Dataset | Percentage |
|---|---|
| Wikipedia Raw Text | 57.6% |
| Public Interview Transcripts | 25.3% |
| TED Talks Transcripts | 17.1% |

Table 1: *Distribution of word counts per dataset.*

The reasons we organize the datasets like this are, on one hand, the large amount of Wikipedia written-form texts are easy to access and Wikipedia texts do great help on capitalization word casing due to its large number of proper nouns; on the other hand, transcripts from public interview and TED Talks enhance the diversity of texts in conversational style. For each dataset, we split the data into 85% training data and 15% validation data. To compare with results in the literature, we test our model on the IWSLT 2011 benchmark dataset [24]. We evaluate punctuation and casing predictions based on precision (P), recall (R) and F$_1$-score (F$_1$).

### 4.2 Data Preprocessing

We split the training, validation and test datasets into sequences of 200 tokens. Each sequence begins with a start-of-sequence (<SOS>) token and ends with an end-of-sequence (<EOS>) token. We use the byte-pair-encoding (BPE) of the Sentence Piece model [25] for tokenizing the word sentence. When a word is tokenized into subword tokens, only the first subword token is labelled for each word. If the addition of subword tokens for a word extends the sequence length beyond 200, we exclude those tokens from the current sequence and start the next sequence from the beginning of that sentence. The padding tokens are appended after the end-of-sequence token to make the sequence length reach 200. To leave a safety margin, we discard sentences with more than 190 tokens. As mentioned in Section 3, the punctuation label set is (O, COMMA, PERIOD, QUESTION) and the casing label set is (O, UPP, CAP, MIX). To generate unformatted text, we strip off all punctuation and capitalization from the target sentence. An example is given in Tabel 2.

### 4.3 Training Details

The model is implemented in PyTorch [26]. We set SentencePiece vocabulary size to 5000. For embedding layer, we use 100-dimension token embeddings. We have carried out extensive experiments with combinations of different numbers of CNN encoder layers, BiLSTM layers, 1D convolution kernel sizes, and other hyper-parameters. The best performer is set up with three CNN encoder layers and two BiLSTM layers, the kernel size of 1D convolution is 3 and the padding of 1D convolution doesn't modify the sequence length (i.e. *same*). The out channels number of 1D convolution is set to the token embedding dimension 100. The hidden layer dimension of BiLSTM layers and LSTM layer are both set to 384. Weights of all layers are initialized by Kaiming uniform initialization [27]. The model is trained using Adam [28] optimizer with a learning rate of 0.002 and weight decay of 2.5e-5. We use ReduceLROnPlateau learning rate scheduler with a factor of 0.8 and a patience of 2. The dropout rate is set to 0.5 for regularization. For the $\alpha$ value in the formulation (6), we explore the values in the range of (0.5-2) and finally find 0.7 to be the optimal value. We use a batch size of 256 and train the model for 30 epochs. The model performing best on validation set is chosen for evaluating test set. It takes about 2 hours to finish the training on a single Nvidia RTX 3090 Ti GPU.

### 4.4 Results and Discussions

We name the proposed model as **CNN-BiLSTM**, the evaluation results of punctuation prediction on IWSLT2011 test set are shown in Table 3. "Overall" refers to the micro-average of scores for all punctuation classes. We compare the proposed CNN-BiLSTM with the existing baselines on the same dataset. The first group of models in Table 3 are non-Transformer models, CNN-2A [14] is purely based on CNN structure. T-LSTM [29] used a unidiretional LSTM model. BLSTM-CRF [30], T-BRNN-pre [17] and Corr-BiRNN [20] are all based on bidirectional RNN model. The second group of models are Transformer based models. Self-attention-word-speech [31] used a full sequence Transformer encoder-decoder model. CT-Transformer [8] only used the encoder part of the Transformer encoder-decoder model structure. In the last group, **CNN-BiLSTM-attention** is a counterpart of CNN-BiLSTM, which adds a scaled dot-product attention [32] layer on top of the unidirectional LSTM layer and before decoder layer.

| Target | | Will AI change our future? Obviously, the answer is yes. |
|---|---|---|
| Unformatted text | | will ai change our future obviously the answer is yes |
| Labels | Punctuation | 0 0 0 0 3 1 0 0 0 2 |
| | Casing | 2 1 0 0 0 2 0 0 0 0 |

Table 2: *Generation of unformatted text and labels. For punctuation labels, mapping 0, 1, 2 and 3 to O, COMMA, PERIOD and QUESTION respectively. For casing, mapping 0, 1, 2 and 3 to O, UPP, CAP and MIX respectively.*

| Model | COMMA | | | PERIOD | | | QUESTION | | | OVERALL | | |
|---|---|---|---|---|---|---|---|---|---|---|---|---|
| | P | R | $F_1$ | P | R | $F_1$ | P | R | $F_1$ | P | R | $F_1$ |
| CNN-2A [14] | 48.1 | 44.5 | 46.2 | 57.6 | 69.0 | 62.8 | 0 | 0 | - | 53.4 | 55.0 | 54.2 |
| T-LSTM [29] | 49.6 | 41.4 | 45.1 | 60.2 | 53.4 | 56.6 | 57.1 | 43.5 | 49.4 | 55.0 | 47.2 | 50.8 |
| BLSTM-CRF [30] | 58.9 | 59.1 | 59.0 | 68.9 | 72.1 | 70.5 | 71.8 | 60.6 | 65.7 | 66.5 | 63.9 | 65.1 |
| T-BRNN-pre [17] | 65.5 | 47.1 | 54.8 | 73.3 | 72.5 | 72.9 | 70.7 | 63.0 | 66.7 | 70.0 | 59.7 | 64.4 |
| Corr-BiRNN [20] | 60.9 | 52.4 | 56.4 | 75.3 | 70.8 | 73.0 | 70.7 | 56.9 | 63.0 | 68.6 | 61.6 | 64.9 |
| Self-attention-word-speech [31] | 67.4 | 61.1 | 64.1 | 82.5 | 77.4 | 79.9 | 80.1 | 70.2 | 74.8 | 76.7 | 69.6 | 72.9 |
| CT-Transformer [8] | 68.8 | 69.8 | 69.3 | 78.4 | 82.1 | 80.2 | 76.0 | 82.6 | 79.2 | 73.7 | 76.0 | 74.9 |
| CNN-BiLSTM-attention | 67.1 | 55.5 | 60.8 | 77.4 | 80.5 | 79.0 | 52.9 | 61.4 | 56.8 | 71.9 | 67.4 | 69.6 |
| CNN-BiLSTM | 71.5 | 54.0 | 61.5 | 76.7 | 82.6 | 79.5 | 73.8 | 70.5 | 72.1 | 74.4 | 67.9 | 71.0 |

Table 3: *Punctuation prediction results on IWSLT2011 test set.*

| Model | UPP-CASE | | | CAP-CASE | | | MIX-CASE | | | OVERALL | | |
|---|---|---|---|---|---|---|---|---|---|---|---|---|
| | P | R | $F_1$ | P | R | $F_1$ | P | R | $F_1$ | P | R | $F_1$ |
| Single-BiRNN [20] | 94.1 | 64.0 | 76.2 | 84.4 | 68.2 | 75.4 | 0 | 0 | - | 88.8 | 75.3 | 81.5 |
| Corr-BiRNN [20] | 93.7 | 60.0 | 73.2 | 82.6 | 71.9 | 76.9 | 0 | 0 | - | 87.2 | 78.2 | 82.4 |
| CNN-BiLSTM-attention | 98.5 | 98.9 | 98.7 | 86.9 | 83.0 | 84.9 | 80.0 | 80.0 | 80.0 | 89.4 | 86.3 | 87.8 |
| CNN-BiLSTM | 98.9 | 98.9 | 98.9 | 85.4 | 84.4 | 84.9 | 80.0 | 80.0 | 80.0 | 88.2 | 87.3 | 87.8 |

Table 4: *Casing prediction results on IWSLT2011 test set. UPP-CASE, CAP-CASE and MIX-CASE denotes all upper case, capitalization case and mixed case respectively.*

| | Example |
|---|---|
| Source | but what are the risks i am not technical i'm not an engineer i don't play one on the internet but i appreciate perspectives on ai from those people who are able to immerse themselves in some of the more technical aspects |
| Gold | But what are the risks? I am not technical. I'm not an engineer. I don't play one on the internet, but I appreciate perspectives on AI from those people who are able to immerse themselves in some of the more technical aspects. |
| CT-Transformer [8] | but what are the risks? i am not technical. i'm not an engineer. i don't play one on the internet, but i appreciate perspectives on ai from those people who are able to immerse themselves in some of the more technical aspects. |
| CNN-BiLSTM | But what are the risks? I am not technical. I'm not an engineer. I don't play one on the internet, but I appreciate perspectives on AI from those people who are able to immerse themselves in some of the more technical aspects. |

Table 5: *Example of CNN-BiLSTM vs. CT-Transformer predictions. CT-Transformer is used as the representative of Transformer based models.*

In comparison to non-Transformer models, our proposed CNN-BiLSTM outperforms all previous models on overall $F_1$-score (9% relative improvement over the best performer in the first group). It is worth noting that the phenomenon where the metrics of COMMA are the worst of all metrics is common to all previous models. This demonstrates that the prediction of COMMA is more challenging and error-prone than that of other punctuation types. Compared to Transformer based models, considering the large model scale and powerful performance of Transformer, it is foreseeable that Transformer based models perform better than the proposed CNN-BiLSTM. But it is surprising that CNN-BiLSTM achieves better results than Transformer based models on some metrics, e.g. the precision of COMMA and the recall of PERIOD. And the overall $F_1$-score of CNN-BiLSTM is only marginally lower than that of Transformer based models. Therefore, it can be said that the results of CNN-BiLSTM are comparable with that of Transformer based models. The CNN-BiLSTM-attention, unexpectedly, achieves lower results than CNN-BiLSTM on the overall F1-score. This demonstrates that the joint learning tasks seemingly do not benefit from the self-attention mechanism.

For casing prediction evaluation, the results on IWSLT2011 test set are shown in Table 4. The proposed CNN-BiLSTM significantly outperforms previous works in terms of $F_1$-score (87.8% versus 82.4%). It is noted that mixed case type was not predicted in Single-BiRNN and Corr-BiRNN. The proposed CNN-BiLSTM is able to predict mixed case type. In Single-BiRNN and Corr-BiRNN, single-letter-word-case (e.g. "I") was listed separately, but we think it is unnecessary. Therefore, we merge this type to all upper case. The CNN-BiLSTM-attention model achieves comparable accuracy to the CNN-BiLSTM on this task.

We use CT-Transformer as the representative of Transformer based models[2]. Then we perform detailed comparisons between CNN-BiLSTM and CT-Transformer in real-time usage. The comparisons are performed under *sherpa-onnx*[3] framework which is based on ONNX [33] format of models. We use Intel Core i5-1035G1 CPU for inference. Table 5 shows the prediction results of the two models on an example paragraph. Both of two models predict punctuations correctly. The CNN-BiLSTM predicts word casing correctly while CT-Transformer does not support casing prediction task. Table 6 shows the comparison results of model size and inference time. The model size means the size of the ONNX format of the models and the inference time is the time cost to finish the prediction of the example paragraph in Table 5. We use the quantized CNN-BiLSTM model in this comparison. As shown in Table 6, the model size of CNN-BiLSTM is only one fortieth of that of CT-Transformer. With respect to inference time, the CNN-BiLSTM is 2.5x faster than CT-Transformer. Considering the much smaller size and faster inference time, the CNN-BiLSTM is more suitable for on-device streaming ASR systems than Transformer based models.

---
[2]The CT-Transformer model is from: `https://github.com/k2-fsa/sherpa-onnx/releases/download/punctuation-models/sherpa-onnx-punct-ct-transformer-zh-en-vocab272727-2024-04-12.tar.bz2`

| Model | Size (MB) | Inference Time (ms) |
|---|---|---|
| CT-Transformer [8] | 280 | 25 (×1.0) |
| CNN-BiLSTM | 7 | 10 (×2.5) |

Table 6: *Detailed comparisons between CNN-BiLSTM and CT-Transformer. The size measures the ONNX model size. The inference time is the time cost to predict the paragraph in Table 5.*

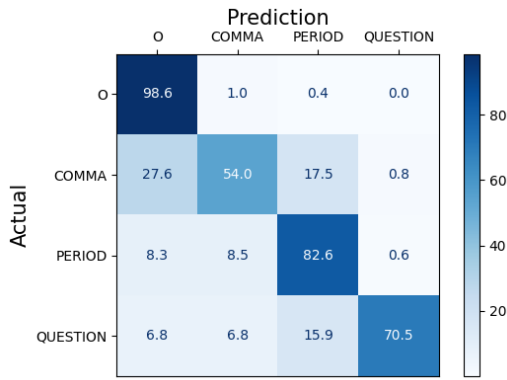

Figure 2: *Punctuation prediction confusion matrix for CNN-BiLSTM on IWSLT2011 test set.*

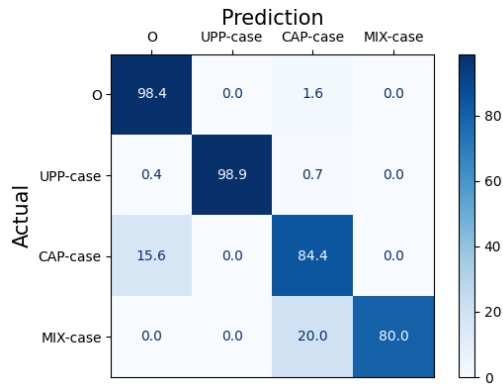

Figure 3: *Casing prediction confusion matrix for CNN-BiLSTM on IWSLT2011 test set.*

We show the confusion matrices of punctuation and casing prediction (in percentage) for CNN-BiLSTM on IWSLT2011 test set in Figure 2 and Figure 3. As shown in Figure 2, COMMA is most frequently mistaken with no punctuation (O). We hypothesize the reason is that commas are often arbitrarily placed in transcribed speech, e.g. a slight pause or a change in pitch from the speaker may be assumed to be a comma. A comparatively high proportion of COMMA is also mistaken with PERIOD. We speculate that the misidentification of commas as periods could be due to the intricate structure of lengthy sentences that contain several commas. These two mistakes of COMMA may explain why the prediction of COMMA is more challenging and error-prone than that of other punctuation types as we discussed above. Another interesting observation is that the proportion of QUESTION being predicted as PERIOD is relatively high. The primary contributing factor to this mistake could be the substantially lower occurrence of question marks compared to periods in training dataset. In Figure 3, we observe that the prediction of CAP-case as all lower case(O) is the biggest mistake of CAP-case. This is most likely caused by the misprediction of considerable number of periods and question marks. Because the tasks of punctuation prediction and casing prediction are interrelated, e.g. the word following a terminal period is always capitalized. Additionally, twenty percent of MIX-case is mistaken with CAP-case. Since mixed case occurs very rarely, so this can be attributed to the scarcity of this type.

## 5. CONCLUSIONS

In this paper, we propose a light-weight and efficient model CNN-BiLSTM designed for the joint prediction of punctuation and word casing. The model uses CNN based encoder to extract information of input words and BiLSTM to capture both past and future contexts for prediction. Experimental results show that CNN-BiLSTM outperforms previous non-Transformer models on $F_1$-scores and achieves comparable results to Transformer based models. Compared with the representative of Transformer based models, the ONNX model size of CNN-BiLSTM is only one fortieth of that of the representative model and CNN-BiLSTM is 2.5 times faster than the representative model regarding inference time. The results demonstrate that CNN-BiLSTM is more suitable for on-device streaming ASR systems than Transformer based models. We have released source code for the research community.

Future work includes increasing the percentage of question marks in training dataset to improve question mark prediction performance, training the model on other languages and integrating the model to ASR system.

---

[3] https://github.com/k2-fsa/sherpa-onnx